# Automatic Evaluation and Uniform Filter Cascades for Inducing $N$-Best Translation Lexicons

I. Dan Melamed
Department of Computer and Information Science
University of Pennsylvania
Philadelphia, PA, 19104
U.S.A.
melamed@unagi.cis.upenn.edu

**Abstract**

**This paper shows how to induce an $N$-best translation lexicon from a bilingual text corpus using statistical properties of the corpus together with four external knowledge sources. The knowledge sources are cast as filters, so that any subset of them can be cascaded in a uniform framework. A new objective evaluation measure is used to compare the quality of lexicons induced with different filter cascades. The best filter cascades improve lexicon quality by up to 137% over the plain vanilla statistical method, and approach human performance. Drastically reducing the size of the training corpus has a much smaller impact on lexicon quality when these knowledge sources are used. This makes it practical to train on small hand-built corpora for language pairs where large bilingual corpora are unavailable. Moreover, three of the four filters prove useful even when used with large training corpora.**

## 1 INTRODUCTION

A machine translation system must be able to choose among possible translations based on context. To do this, it usually relies on a translation lexicon that contains a number of possible translations for each word. *$N$-best translation lexicons* contain up to $N$ candidate translations for each word, ordered from most probable to least probable, sometimes specifying *a priori* probabilities or likelihood scores.

Existing automatic methods for constructing $N$-best translation lexicons rely on the availability of large training corpora of parallel texts in the source and target languages. For some methods, the corpora must also be aligned by sentence [Bro93, Gal91a]. Unfortunately, such training corpora are available for only a handful of language pairs, and the cost to create enough training data manually for new language pairs is very high.

This paper presents

1. a new automatic evaluation method for $N$-best translation lexicons,
2. a filter-based approach for enhancing statistical translation models with non-statistical sources of information,
3. four sources of information that can drastically reduce the necessary amount of training material.

The evaluation method uses a simple objective criterion rather than relying on subjective human judges. It allows many experiments to be run without concern about the cost, availability and reliability of human evaluators.

The filter-based approach is designed to identify likely (source word, target word)[1] pairs, using a statistical decision procedure. Candidate word pairs are drawn from a corpus of aligned sentences: (S, T) is a candidate iff T appears in the translation of a sentence containing S. In the simplest case, the decision procedure considers all candidates for inclusion in the lexicon; but the new framework allows a cascade of non-statistical filters to remove inappropriate pairs from consideration.

Each filter is based on a particular knowledge source, and can be placed into the cascade independently of the others. The knowledge sources investigated here are:

- part of speech information,
- machine-readable bilingual dictionaries (MRBDs),
- cognate heuristics, and
- word alignment heuristics.

[Bro94] investigated the statistical use of MRBDs, though not as filters. The other three knowledge sources have not previously been used for the task of inducing translation lexicons.

The filter-based framework, together with the fully automatic evaluation method, allows easy investigation of the relative efficacy of cascades of each of the subsets of these four filters. As will be shown below, some filter cascades sift candidate word pairs so well that training corpora small enough to be hand-built can be used to induce more accurate translation lexicons than those induced from a much larger training corpus without such filters. In one evaluation, a training corpus of 500 sentence pairs processed with these knowledge sources achieved a precision of 0.54, while a training corpus of 100,000 training pairs alone achieved a precision of only 0.45. Such improvements could not be previously obtained, because

- These knowledge sources have not been used together for this task before.
- There was no way to uniformly combine the different kinds of filters.
- There was no way to objectively judge lexicon precision.

Table 1 provides a qualitative demonstration of how a lexicon entry gradually improves as more filters are applied. The table contains actual entries for the French source word "premier," from 7-best lexicons that were induced from 5000 pairs of training sentences, using different filter cascades. The baseline lexicon, induced with no filters, contains correct translations only in the first and sixth positions. The Cognate Filter disallows all candidate translations of French "premier" whenever the English cognate "premier" appears in the target English sentence. This causes English "premier" to move up to second position. The Part-of-Speech Filter realizes that "premier" can only be an adjective in French, whereas in the English Hansards it is mostly used as a noun. So, it throws out that pairing, along with several other English noun candidates, allowing "first" to move up to third position. The POS and Cognate filters reduce noise better together than separately. More of the incorrect translations are filtered out in the "POS & COG" column, making room for "foremost." Finally, the MRBD Filter narrows the list down to just the three translations of French "premier" that are appropriate in the Hansard sublanguage.

[1] Punctuation, numbers, *etc.* also count as words.

Table 1: entries for French "premier" in 7-best lexicons generated using different cascades of filters

| Entry # | No Filters | COG Filter | POS Filter | COG & POS | COG, POS & MRBD |
|---|---|---|---|---|---|
| 1 | **prime** | **prime** | **prime** | **prime** | **prime** |
| 2 | minister | **premier** | direct | direct | **first** |
| 3 | **premier** | direct | **first** | **first** | **foremost** |
| 4 | direct | Speaker | supplementary | former | |
| 5 | question | Mr. | former | friendly | |
| 6 | **first** | my | friendly | **foremost** | |
| 7 | Speaker | **first** | reaffirm | echo | |

## 2 EXPERIMENTAL FRAMEWORK

All translation lexicons discussed in this paper were created and evaluated using the procedure in Figure 1. First, candidate translations were generated for each pair of aligned training sentences, by taking a simple cross-product of the words. Next, the candidate translations from each pair of training sentences were passed through a cascade of filters. The remaining candidate translations from all training sentence pairs were pooled together and fed into a fixed decision procedure. The output of the decision procedure was a model of word correspondences between the two halves of the training corpus — a translation lexicon. Each filter combination resulted in a different model. All the models were compared in terms of how well they represented a held-out test set. The evaluation was performed objectively and automatically using Bitext-Based Lexicon Evaluation (BiBLE, described below). BiBLE assigned a score for each model, and these scores were used to compare the effectiveness of various filter cascades.

As shown in Figure 1, the only independent variable in the framework is the cascade of filters used on the translation candidates generated by each sentence pair, while the only dependent variable is a numerical score. Since the filters only serve to remove certain translation candidates, any number of filters can be used in sequence. This arrangement allows for fair comparison of different filter combinations.

## 3 BITEXT-BASED LEXICON EVALUATION (BiBLE)

Translation lexicon quality has traditionally been measured on two axes: precision and recall. *Recall* is the fraction of the source language's vocabulary that appears in the lexicon. *Precision* is the fraction of lexicon entries that are correct. While the true size of the source vocabulary is usually unknown, recall can be estimated using a representative text sample by computing the fraction of words in the text that also appear in the lexicon. Measuring precision is much more difficult, because it is unclear what a "correct" lexicon entry is — different translations are appropriate for different contexts, and, in most cases, more than one translation is correct. This is why evaluation of translation has eluded automation efforts until now.

The large number of quantitative lexicon evaluations required for the present study made it infeasible to rely on evaluation by human judges. The only existing automatic lexicon evaluation method that I am aware of is the perplexity comparisons used by Brown *et al.* in the framework of their Model 1 [Bro93]. Lexicon perplexity indicates how "sure" a translation lexicon is about its contents. It does not, however, directly measure the quality of those contents.

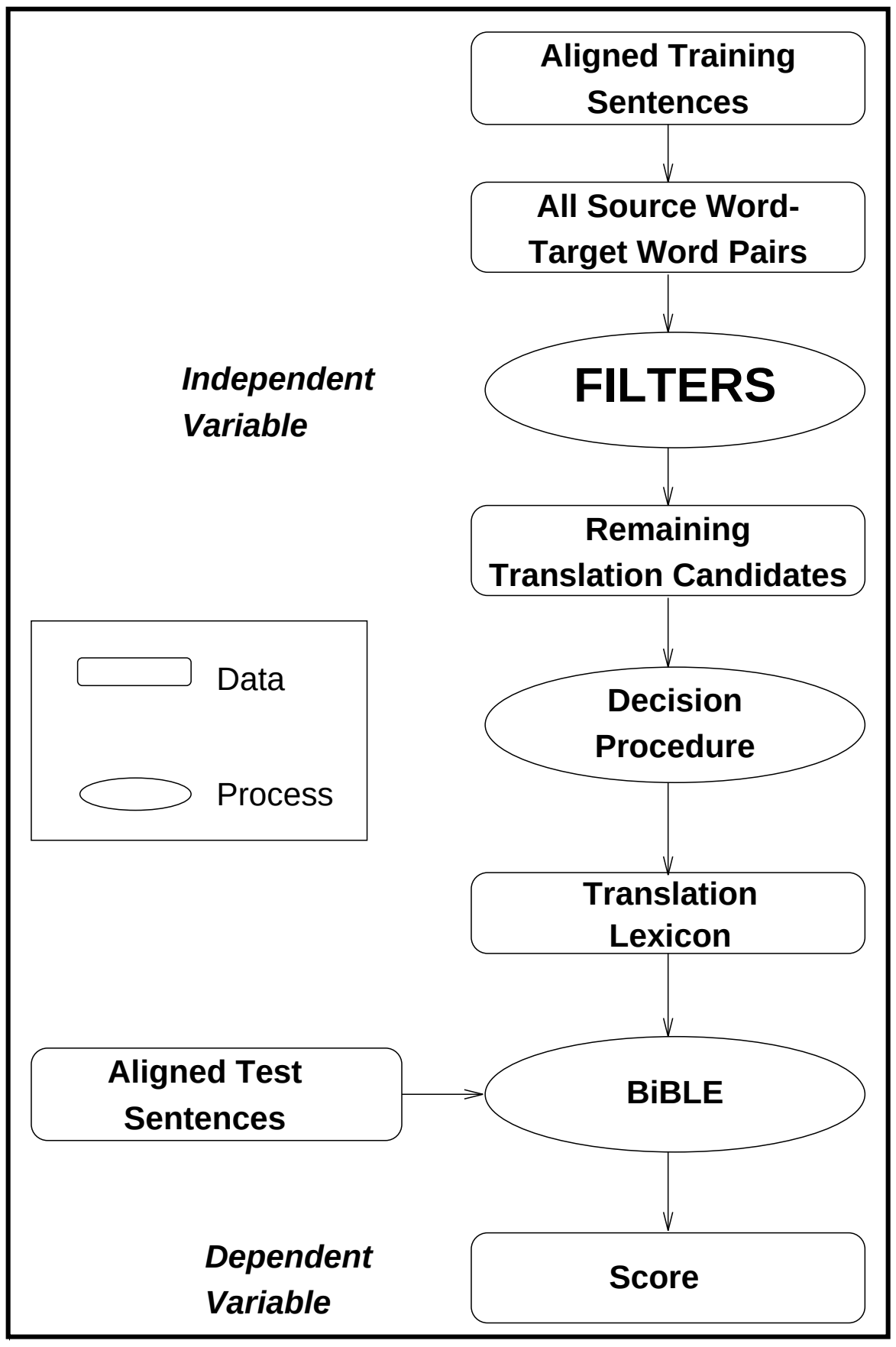

Figure 1: Uniform Framework for Data Filters

BiBLE is a family of algorithms, based on the observation that *translation pairs*[2] tend to appear in corresponding sentences in an aligned bilingual text corpus (a *bitext*). Given a test set of aligned sentences, a better translation lexicon will contain a higher fraction of the (source word, target word) pairs in those sentences. This fraction can be computed either by token or by type, depending on the application. If only the words in the lexicon are considered, BiBLE gives an estimate of precision. If all the words in the text are considered, then BiBLE measures percent correct. Note that BiBLE treats the test sentences as bags of words; the order of those words is not considered. The greater the overlap between the vocabulary of the test bitext and the vocabulary of the lexicon being evaluated, the more confidence can be placed in the BiBLE score.

The BiBLE approach is suitable for many different evaluation tasks. Besides comparing different lexicons on different scales, BiBLE can be used to compare different parts of one lexicon that has been partitioned using some characteristic of its entries. For example, the quality of a lexicon's noun entries can be compared to the quality of its adjective entries; the quality of its entries for frequent words can be compared to the quality of its entries for rare words. Likewise, separate evaluations can be performed for each $k$, $1 \leq k \leq N$, in $N$-best lexicons.

Figure 2 shows the outline of a BiBLE algorithm for evaluating precision of $N$-best translation lexicons. The $k$th cumulative hit rate for a source word S is the fraction of test sentences containing

[2] A "translation pair" is a source word and a target word that are translations of each other.

Figure 2: A Bitext-Based Lexicon Evaluation (BiBLE) algorithm for precision of $N$-best lexicons — Percent correct can be evaluated instead of precision by switching lines 3 and 4.

Input:

1. translation lexicon with up to $N$ translations for each word
2. aligned test bitext

Algorithm:

```
 1  FOR EACH pair of aligned test sentences
 2      FOR EACH word S in the source sentence
 3          IF S is in the lexicon
 4              frq(S) += 1
 5              k = 0
 6              found = false
 7              DO UNTIL found OR k = N
 8                  k += 1
 9                  IF S's kth translation T is in the target sentence
10                      delete T from target sentence
11                      HitCount[S,k] += 1
12                      found = true

13  FOR EACH word S in the source vocabulary
14      IF frq(S) > 0
15          FOR k = 1 TO N
16              HitRate[k] += HitCount[S,k] / frq(S)

17  CumulativeHitRate[0] = 0
18  FOR k = 1 TO N
19      CumulativeHitRate[k] = CumulativeHitRate[k-1] + HitRate[k]
```

Output:

```
CumulativeHitRate[1..N]
```

S whose translations contain one of the $k$ best translations of S in the lexicon. For each $k$, the $k$th cumulative hit rates are averaged over all the source words in the lexicon, counting words by type. This yields $N$ average cumulative hit rates for the lexicon as a whole.

In this study, the average is computed by type and not by token, because translations for the most frequent words are easy to estimate using any reasonable statistical decision procedure, even without any extra information. Token-based evaluation scores would be misleadingly inflated with very little variation. Computing hit rates for each word separately and then taking an unweighted average ensures that a correct translation of a common source word does not contribute more to the score than correct translations of rare words. The evaluation is uniform over the whole lexicon.

BiBLE evaluation is quite harsh, because many translations are not word for word in real bitexts. To put BiBLE scores reported here into proper perspective, human performance was evaluated on a similar task. The 1994 ARPA-sponsored machine translation evaluation effort generated two independent English translations of one hundred French newspaper texts [Whi93]. I hand-aligned each pair of translations by paragraph; most paragraphs contained between one and four sentences. For each pair of translations, the fraction of times (by type) that identical words were used in corresponding paragraphs was computed. The average of these 100 fractions was 0.6182 with a standard deviation of 0.0647. This is a liberal estimate of the upper bound on the internal consistency of BiBLE test sets. Scores for sentence-based comparisons will always be lower than scores for paragraph-based comparisons, because there will be fewer spurious "hits." To confirm this, an independent second translation of 50 French Hansard sentences was commissioned. The translation scored 0.57 on this test.

# 4 EXPERIMENTS

A bilingual text corpus of Canadian parliamentary proceedings ("Hansards") was aligned by sentence using the method presented in [Gal91b]. From the resulting aligned corpus, this study used only sentence pairs that were aligned one to one, and then only when they were less than 16 words long and aligned with high confidence. Morphological variants in these sentences were stemmed to a canonical form. Fifteen thousand sentence pairs were randomly selected and reserved for testing; one hundred thousand were used for training.

The independent variable in the experiments was a varying combination of four different filters, used with six different sizes of training corpora. These four filters fall into three categories: predicate filters, oracle filters and alignment filters. A *predicate filter* is one where the candidate translation pair (S, T) must satisfy some predicate in order to pass the filter. Various predicate filters are discussed in [Wu94]. An *oracle filter* is useful when a list of likely translation pairs is available *a priori*. Then, if the translation pair (S, T) occurs in this oracle list, it is reasonable to filter out all other translation pairs involving S or T in the same sentence pair. An *alignment filter* is based on the relative positions of S and T in their respective texts[Dag93].

The decision procedure used to select lexicon entries from the multiset of candidate translation pairs is a variation of the method presented in [Gal91a]. [Dun93] found binomial log-likelihood ratios to be relatively accurate when dealing with rare tokens. This statistic was used to estimate dependencies between all co-occuring (source word, target word) pairs. For each source word S, target words were ranked by their dependence with S. The top $N$ target words in the rank-ordering for S formed the entry for S in the $N$-best lexicon. In other words, the relative magnitude of dependence between S and its candidate translations was used as a maximum likelihood estimator of the translations of S.

### 4.1 Part of Speech Filter

The POS Filter is a predicate filter. It is based on the idea that word pairs that are good translations of each other are likely to be the same parts of speech in their respective languages. For example, a noun in one language is very unlikely to be translated as a verb in another language. Therefore, candidate translation pairs involving different parts of speech should be filtered out.

This heuristic should not be taken too far, however, in light of the imperfection of today's tagging technology. For instance, particles are often confused with prepositions and adjectives with past participles. These considerations are further complicated by the differences in the tag sets used by taggers for different languages. To maximize the filter's effectiveness, tag sets must be remapped to a more general common tag set, which ignores many of the language-specific details. Otherwise, correct translation pairs would be filtered out because of superficial differences like tense and capitalization.

The different ways to remap different tag sets into a more general common tag set represent a number of design decisions. Fortunately, BiBLE provided an objective criterion for tag set design, and a fast evaluation method. The English half of the corpus was tagged using Brill's transformation-based tagger [Bri92]. The French half was kindly tagged by George Foster of CITI. Then, BiBLE was used to select among several possible generalizations of the two tag sets. The resulting optimal tag set is shown in Table 2.

Table 2: optimal common tag set for POS Filter

| Tag | Meaning | Matches |
|---|---|---|
| CD | number | CD |
| CJ | conjunction | CJ |
| D | determiner | D |
| EOP | end of phrase marker (",", ";", *etc.*) | EOP |
| EOS | end of sentence marker (".", "?", *etc.*) | EOS |
| IN | preposition or particle | IN |
| J | adjective | J, VBG, VBN |
| N | noun (including "$") | N, NP |
| NP | proper noun | NP, N |
| P | pronoun | P |
| R | adverb | R |
| SCM | subordinate clause marker (quotes, brackets, *etc.*) | SCM |
| UH | interjection | UH |
| V | verb | V |
| VBG | present participle | VBG, J, VBN |
| VBN | past participle | VBN, J, VBG |

### 4.2 Machine-Readable Bilingual Dictionary (MRBD)

An oracle list of 53363 one-to-one translation pairs was extracted from the Collins French-English MRBD [Cou91]. Whenever a candidate translation pair (S,T) appeared in the list of translations extracted from the MRBD, the filter removed all word pairs (S, not T) and (not S, T) that occurred in the same sentence pair.

The MRBD Filter is an oracle filter. It is based on the assumption that if a candidate translation pair (S,T) appears in an oracle list of likely translations, then T is the correct translation of S in their sentence pair, and there are no other translations of S or T in that sentence pair. This assumption is stronger than the one made by Brown *et al.* [Bro94], where the MRBD was treated as data and not as an oracle. Brown *et al.* allowed the training data to override information gleaned from the MRBD. The attitude of the present study is "Don't guess when you know." This attitude may be less appropriate when there is less of an overlap between the vocabulary of the MRBD and the vocabulary of the training bitext, as when dealing with technical text or with a very small MRBD.

The presented framework can be used as a method of enhancing an MRBD. Merging an MRBD with an $N$-best translation lexicon induced using the MRBD Filter will result in an MRBD with more entries that are relevant to the sublanguage of the training bitext. All the relevant entries will be rank ordered for appropriateness.

## 4.3 Cognate Filter

A Cognate Filter is another kind of oracle filter. It is based on the simple heuristic that if a source word S is a cognate of some target word T, then T is the correct translation of S in their sentence pair, and there are no other translations of S or T in that sentence pair. Of course, identical words can mean different things in different languages. The cognate heuristic fails when dealing with such *faux amis* [Mac94]. Fortunately, between French and English, true cognates occur far more frequently than *faux amis*.

There are many possible notions of what a cognate is. Simard et al. used the criterion that the first four characters must be identical for alphabetic tokens to be considered cognates [Sim92]. Unfortunately, this criterion produces false negatives for pairs like "government" and "gouvernement", and false positives for words with a great difference in length, like "conseil" and "conservative." I used an approximate string matching algorithm to capture a more general notion of cognateness. Whether a pair of words is considered a cognate pair depends on the ratio of the length of their longest (not necessarily contiguous) common subsequence to the length of the longer word. This is called the Longest Common Subsequence Ratio (LCSR). For example, "gouvernement," which is 12 letters long, has 10 letters that appear in the same order in "government." So, the LCSR for these two words is 10/12. On the other hand, the LCSR for "conseil" and "conservative" is only 6/12. The only remaining question was what minimum LCSR value should indicate that two words are cognates. This question was easy to answer using BiBLE. BiBLE scores were maximized for lexicons using the Cognate Filter when a LCSR cut-off of 0.58 was used. The Wilcoxon signed ranks test found the difference between BiBLE scores for lexicons produced with this LCSR cut-off and for lexicons produced with the criterion used in [Sim92] to be statistically significant at $\alpha = 0.01$. The longest common subsequence between two words can be computed as a special case of their edit distance, in time proportional to the product of their lengths[Wag74].[3]

## 4.4 Word Alignment Filter

Languages with a similar syntax tend to express ideas in similar order. The translation of a word occurring at the end of a French sentence is likely to occur towards the end of the English translation. In general, lines drawn between corresponding lexemes in a French sentence and its

[3] Due to time constraints, I report results for a greedy approximation of the LCSR. A proper implementation might perform even better.

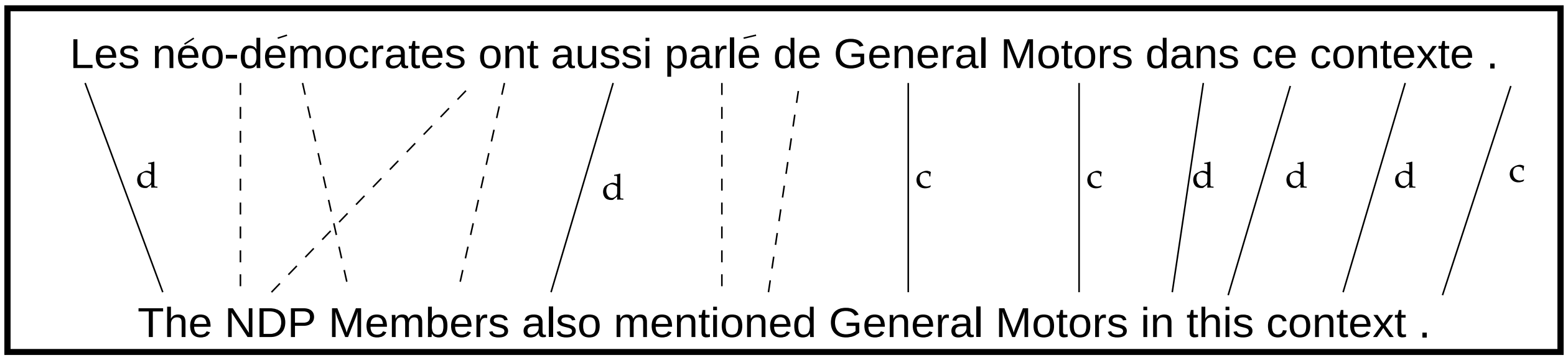

Figure 3: Word Alignment Filter — Partitioning loci marked “d” are translation pairs found in the MRBD, while those marked “c” are cognates. The remaining uncertainties are marked with dashed lines. The Word Alignment Filter removes from consideration candidate translation pairs like (ont, mentioned) which would cross the partition created by (aussi, also).

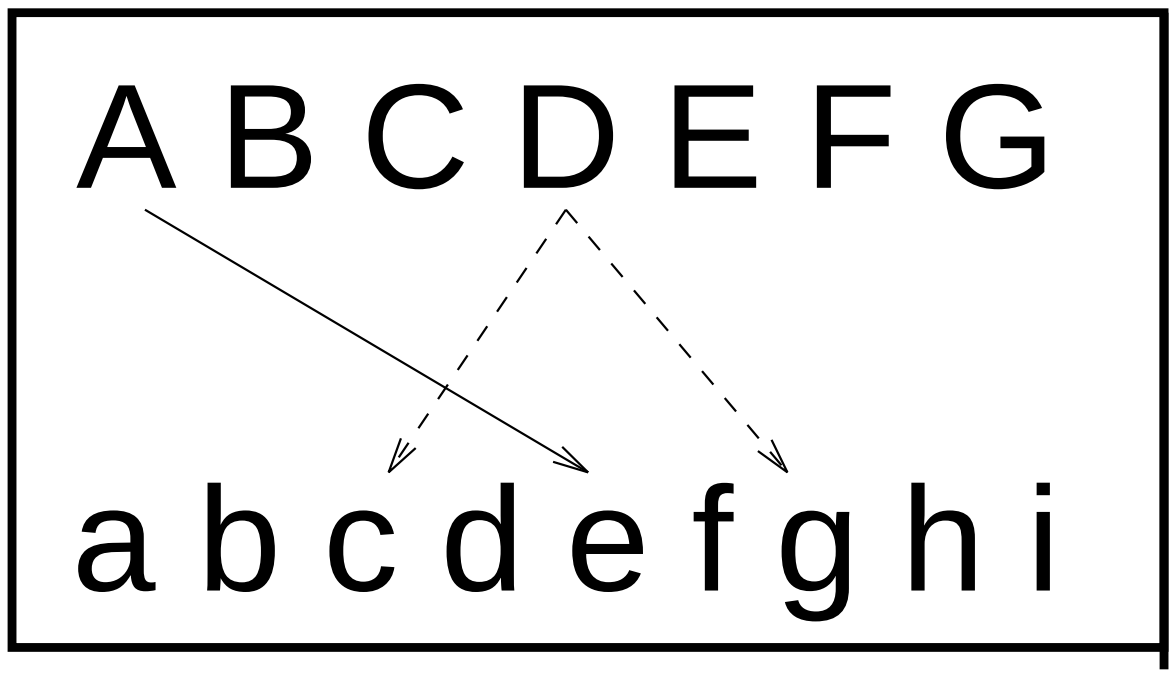

Figure 4: One of the heuristics used in the Word Alignment Filter — Crossing partitions are minimized by aligning D with g rather than with c.

English translation will be mostly parallel. This idea of translation alignment was central to the machine translation method pioneered at IBM [Bro93].

The Word Alignment Filter exploits this observation, as illustrated in Figure 3. If word T in a target sentence is the translation of word S in the corresponding source sentence, then words occurring before S in the source sentence will likely correspond to words occurring before T in the target sentence. Likewise, words occurring after S in the source sentence will likely translate to words occurring after T in the target sentence. So S and T can be used as loci for partitioning the source and target sentences into two shorter pairs of corresponding word strings. Each such partition reduces the number of candidate translations from each sentence pair by approximately a factor of two — an excellent noise filter for the decision procedure.

The Word Alignment Filter is particularly useful when oracle lists are available to identify a large number of translation pairs that can be used to partition sentences. Using a LCSR cut-off of 0.58 (optimized using BiBLE, of course), cognates were found for 23% of the source tokens in the training corpus (counting punctuation). 47% of the source tokens were found in the MRBD. Although there was some overlap, an average of 63% of the words in each sentence were paired up with a cognate or with a translation found in the MRBD, leaving few candidate translations for the remaining 37%.

The oracles lists often supplied more than one match per word. For instance, several determiners or prepositions in the French sentence often matched the same word in the English sentence. When this happened, the current implementation of the Word Alignment Filter used several heuristics to choose at most one partitioning locus per word. For example, one heuristic says that the order of ideas in a sentence is not likely to change during translation. So, it aimed to minimize crossing partitions, as shown in Figure 4. If word A matches word e, and word D matches words c and g, then D is paired with g, so that when the sentences are written one above the other, the lines connecting the matching words do not cross. Between French and English, this heuristic works quite well, except when it comes to the order between nouns and adjectives.

### 4.5 Evaluation

Table 1 is unusual: It is atypical for more than two of the filters studied here to incrementally improve one lexicon entry. Most lexicon entries are improved by just one or two filters, after which more filtering gives no significant benefit. However, each filter improves a large number of different entries. Two more examples of the benefits of different filter cascades are given in Tables 3 and 4.

Table 3: lexicon entries for French "grand" in 7-best lexicons generated with different filters — The baseline lexicon has correct entries only for the most likely translation and for the second most likely translation. The POS Filter throws out nouns and pronouns, and makes room for "high" and "vast." The Word Alignment Filter removes enough noise to capture "high," "vast," "giant," and "extensive" all at once.

| Entry # | No Filters (baseline) | POS Filter | Cognate & POS Filters with Word Alignment |
|---|---|---|---|
| 1 | **great** | **great** | **great** |
| 2 | **large** | **large** | **large** |
| 3 | corporation | **high** | **high** |
| 4 | 's | developmental | **vast** |
| 5 | more | humble | humble |
| 6 | one | undeniable | **giant** |
| 7 | developmental | **vast** | **extensive** |

Table 4: lexicon entries for French "parti" in 7-best lexicons generated with different filters — Only the most likely translation and the fourth most likely translation in the baseline lexicon are appropriate. The Cognate Filter allows the fourth item, a cognate, to percolate up to second place, and makes room for "two-party" in sixth place.

| Entry # | No Filters | Cognate Filter |
|---|---|---|
| 1 | **Party** | **Party** |
| 2 | Liberal | **party** |
| 3 | Democratic | stretch |
| 4 | **party** | handbook |
| 5 | Conservative | espouse |
| 6 | new | **two-party** |
| 7 | the | between |

Figures 5 and 6 show mean BiBLE scores for precision of the best translations in lexicons induced with various cascades of the four filters discussed. Assuming that BiBLE scores are normally distributed, 95% confidence intervals were estimated for each score, using ten mutually exclusive training sets of each size. All the confidence intervals were narrower than one percentage point at 500 pairs of training sentences, and narrower than half of one percentage point at 2000 pairs. Therefore, BiBLE score differences displayed in Figures 5 and 6 are quite reliable.

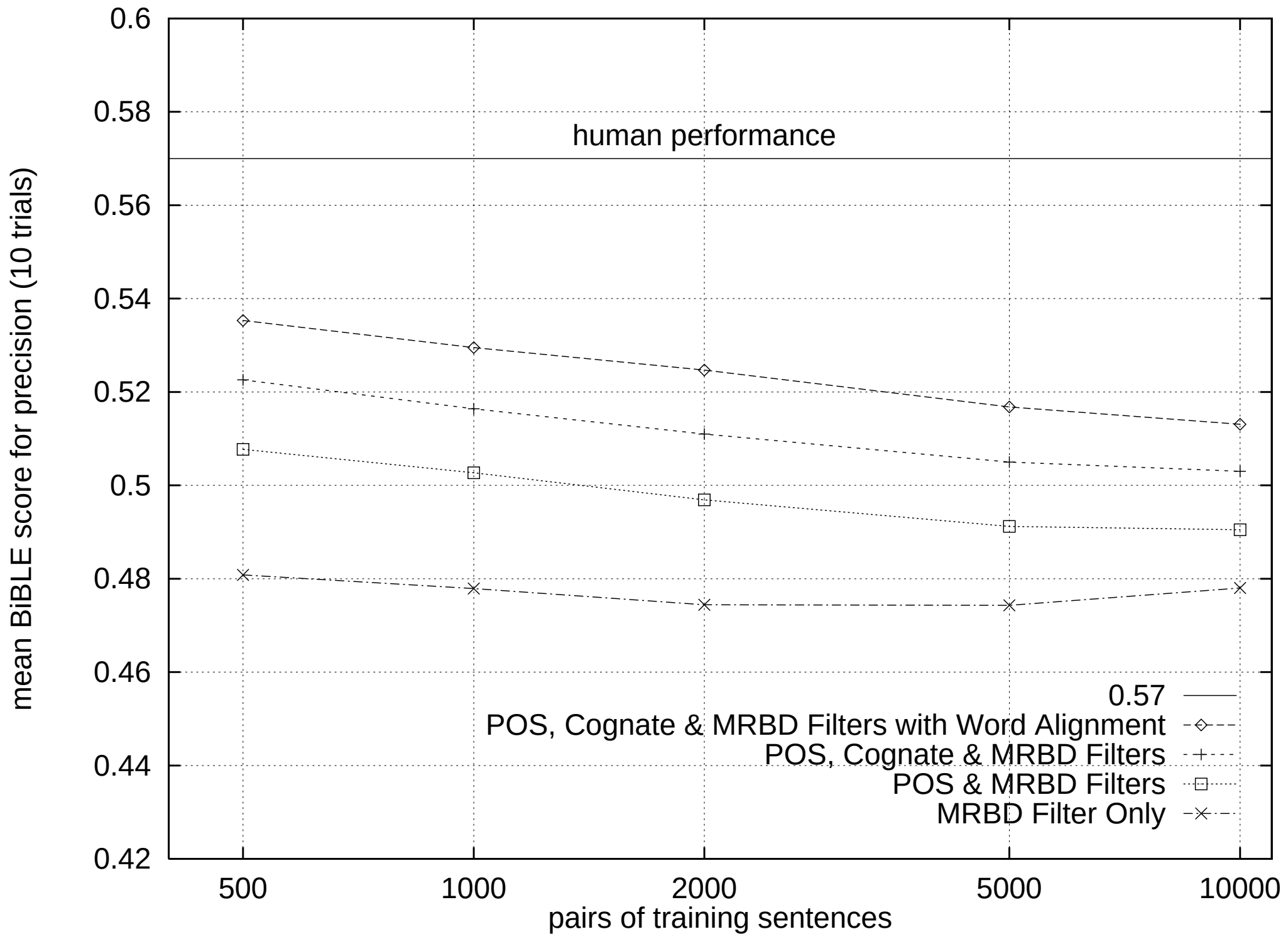

Figure 5: The large MRBD resulted in the most useful filter for this pair of languages. The scores for the cascade of all the filters (the highest curve) are close to the human performance of 0.57.

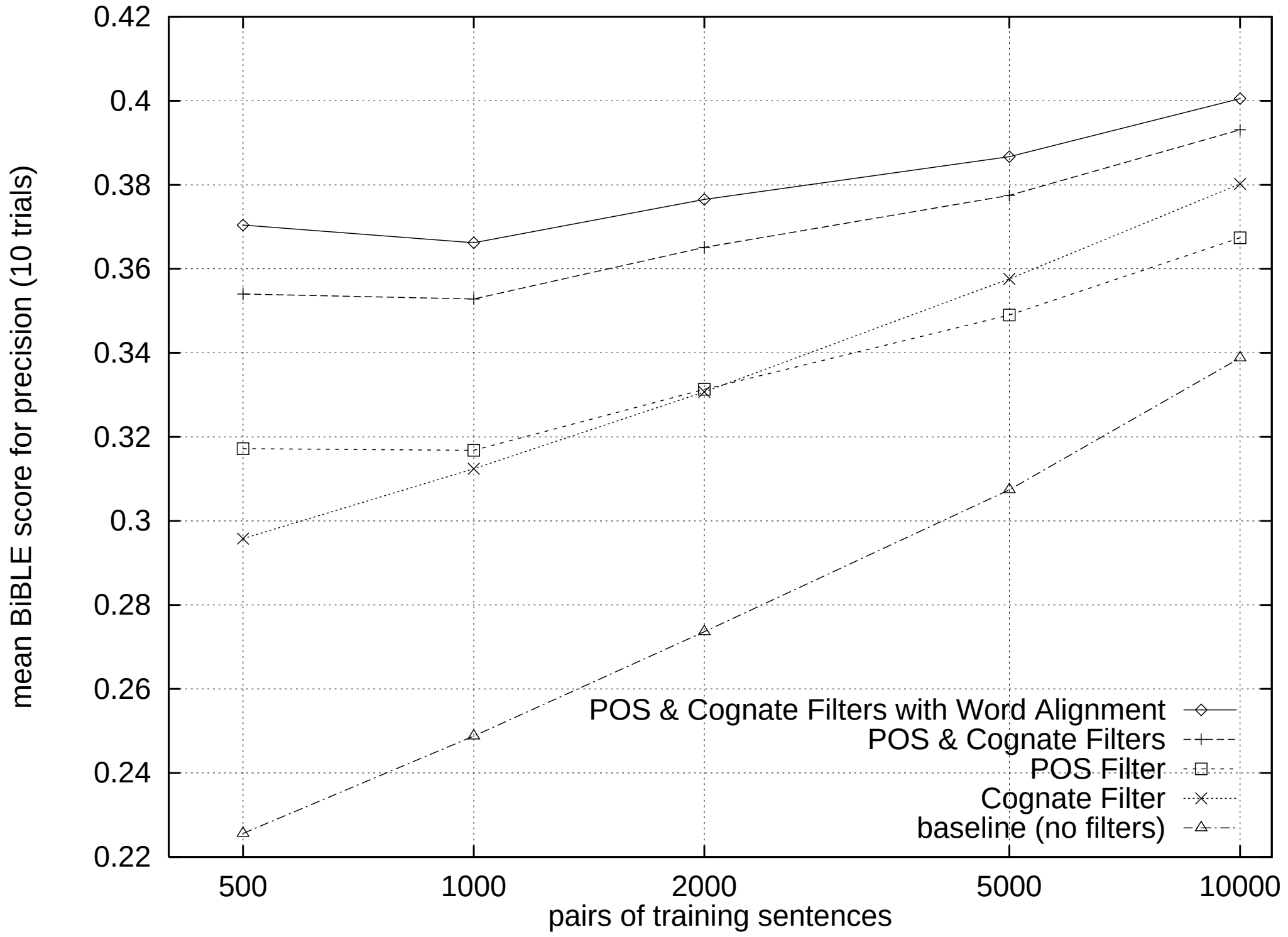

Figure 6: Each filter contributes to an improvement in BiBLE scores.

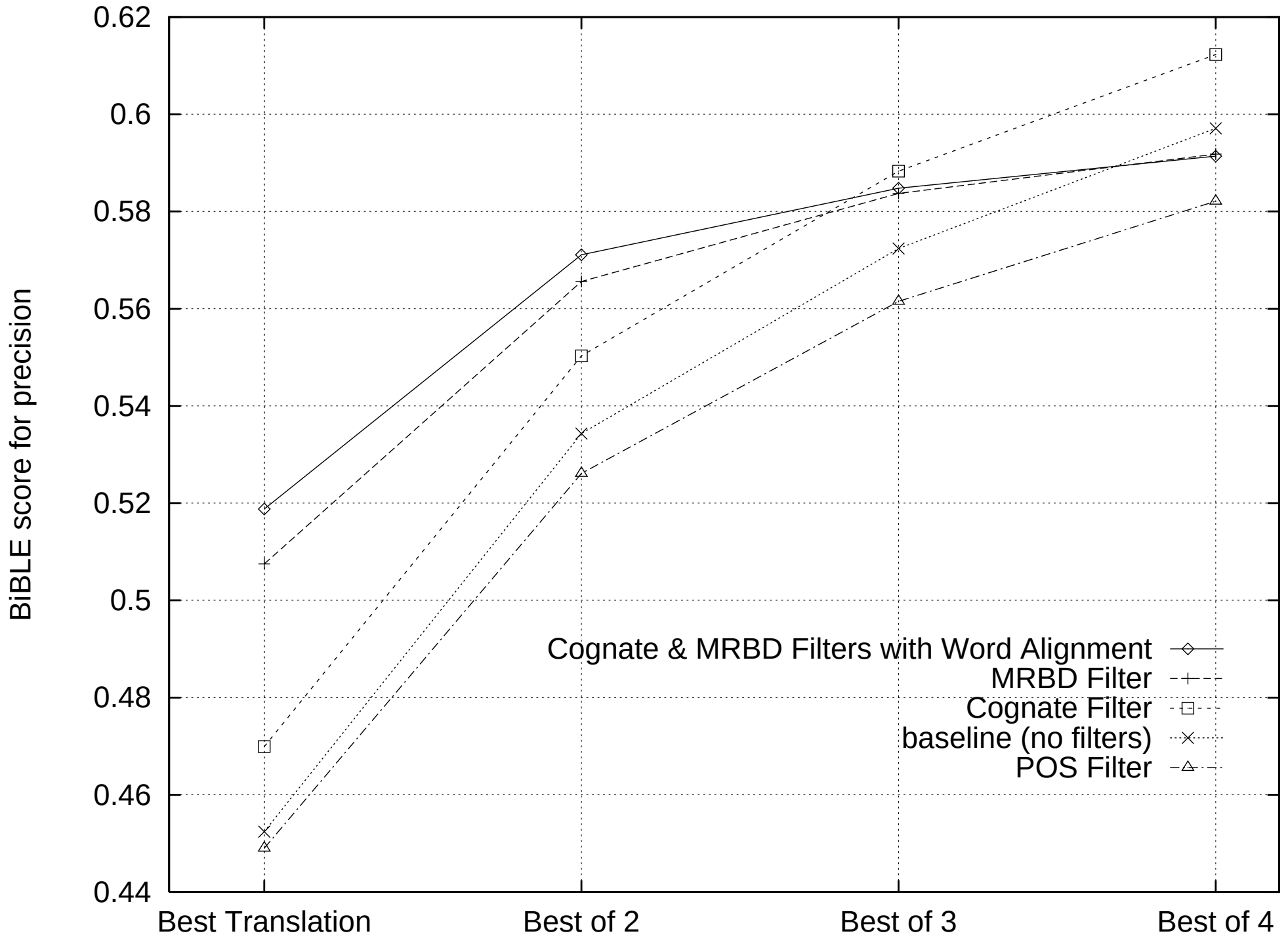

Figure 7: Some useful filter cascades for training corpora as large as 100000 sentence pairs. The Cognate Filter by itself achieves the best precision for the best-of-$N$ translations, when $N > 2$. The POS Filter only degrades precision for large training corpora.

The upper bound on performance for this task is plotted at 0.57 (see end of Section 3). The better filter cascade produce lexicons whose precision comes close to this mark. The best cascades are up to 137% more precise than the baseline model. The large MRBD resulted in the most useful filter for this pair of languages. Future research will look into why the MRBD's contribution to lexicon precision decreases with more training data.

Figure 7 shows the relative performance of selected filters when the entire training set of one hundred thousand sentences is used. All the presented filters, except the POS Filter, improve performance even when a large training corpus is available. Evidently, some information that is useful for inducing translation lexicons cannot be inferred from any amount of training data using only simple statistical methods. The best precision for the single best translation is achieved by a cascade of the MRBD, Cognate and Word Alignment Filters. To maximize precision for the best of three or more translations, only the Cognate Filter should be used.

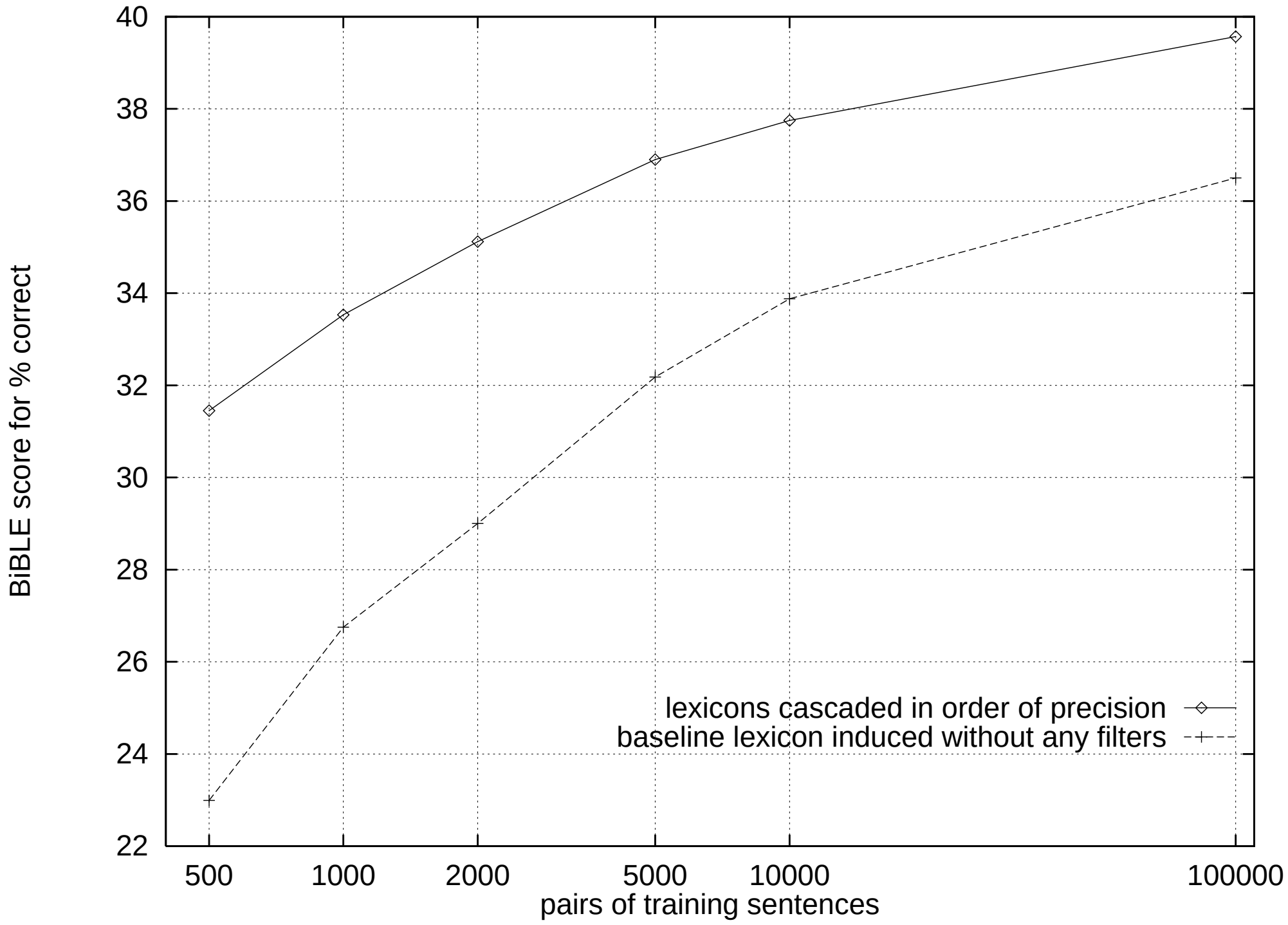

Figure 8: Percent Correct by token – Data filters can improve these scores by more than 35%.

# 5 APPLICATION TO MACHINE-ASSISTED TRANSLATION

A machine translation system should not only translate with high precision, but it should also have good coverage of the source language. So, the product of recall and precision, *percent correct*, is a good indication of a lexicon's suitability for use with such a system. This statistic actually represents the percentage of words in the target test corpus that would be correctly translated from the source, if the lexicon were used as a simple map, Therefore, if the lexicon is to be used as part of a machine-assisted translation system, then the percent correct score will be inversely proportional to the required post-editing time.

A simple strategy was adopted to demonstrate the practical utility of filters presented in this paper. First, the most precise filter cascade was selected by looking at Figure 5. Translations were found for all words in the test source text that had entries in the lexicon induced using that cascade. Then the second most precise filter cascade was selected. Words that the most precise lexicon "didn't know about," which were found in the second most precise lexicon, were translated next. All the other available lexicons were cascaded this way, in the order of their apparent precision, down to the baseline lexicon. This "cascaded back-off" strategy maintained the recall of the baseline lexicon, while taking advantage of the higher precision produced by various filter cascades.

Although more sophisticated translation strategies are certainly possible, BiBLE percent correct scores for cascaded lexicons suffice to test the utility of data filters for machine translation. The results in Figure 8 indicate that the filters described in this paper can be used to improve the performance of lexical transfer models by more than 35%.

## 6 CONCLUSIONS

The research presented here makes several contributions to research in machine translation and related fields:

- a uniform framework for combining various data filters with statistical methods for inducing $N$-best translation lexicons,
- an automatic evaluation method for translation lexicons which obviates the need for labor-intensive subjective evaluation by human judges,
- four different ways to improve statistical translation models,
- a demonstration of how tiny training corpora can be enhanced with non-statistical knowledge sources to induce better lexicons than unenhanced training corpora many times the size.

The effectiveness of different data filters for inducing translation lexicons crucially depends on the particular pair of languages under consideration. *Cognates* are more common, and therefore more useful, in languages which are more closely related. For example, one would expect to find more cognates between Russian and Ukrainian than between French and English. The implementation of a *part of speech* filter for a given pair of languages depends on the availability of part of speech taggers for both languages, where the two taggers have a small common tag set. The effectiveness of oracle filters based on *MRBDs* will depend on the extent to which the vocabulary of the MRBD intersects with the vocabulary of the training text. This, in turn, depends partly on the size of the MRBD. Filters based on *word alignment* patterns will only be as good as the model of typical word alignments between the pair of languages in question. For languages with very similar syntax, a linear model will suffice. Higher order models will be required for a pair of languages like English and Japanese.

For the case of French and English, each of the presented filters makes a significant improvement over the baseline model. Taken together, the filters produce models which approach human performance. These conclusions could not have been drawn without a uniform framework for filter comparison or without a technique for automatic evaluation. An automatic evaluation technique such as BiBLE should be used to gauge the effectiveness of any MT system which has a lexical transfer component. BiBLE's objective criterion is quite simple, with the drawback that it gives no indication of what kinds of errors exist in the lexicon being evaluated. Even so, given a test corpus of a reasonable size, it can detect very small differences in quality between two $N$-best translation lexicons. For example, BiBLE evaluations were used to find the precise optimum value for the LCSR cut-off in the Cognate Filter. BiBLE also helped to select the optimum tag set for the POS Filter. This kind of automatic quality control is indispensable for an engineering approach to better machine translation.

## 7 ACKNOWLEDGEMENTS

I am deeply grateful to George Foster for POS-tagging the French half of my text corpus, to Matthew Stone for providing a second translation of some Hansard text, and to the following people for valuable advice and discussions: Ken Church, Michael Collins, Jason Eisner, George Foster, Mark Liberman, Mitch Marcus, Adwait Ratnaparkhi, Jeff Reynar, Henry Thompson, David Yarowsky, and four anonymous reviewers. This research was partially supported by ARO Contract DAAL03-89-C0031 and by ARPA Contract N6600194-c6043.